\documentclass{article}

\usepackage[preprint]{neurips_2024}
\usepackage[pdftex]{graphicx}
\usepackage{wrapfig}
\usepackage{subcaption}

\usepackage[utf8]{inputenc} % allow utf-8 input
\usepackage[T1]{fontenc}    % use 8-bit T1 fonts
\usepackage{hyperref}       % hyperlinks
\usepackage{url}            % simple URL typesetting
\usepackage{booktabs}       % professional-quality tables
\usepackage{amsfonts}       % blackboard math symbols
\usepackage{nicefrac}       % compact symbols for 1/2, etc.
\usepackage{microtype}      % microtypography
\usepackage{xcolor}         % colors

\title{Optimizing Waste Management with Advanced Object Detection for Garbage Classification}

\author{%
  Everest Z. Kuang \\
  Shaker High School\\
  Watervliet, NY\\
  \texttt{everestkuang9@gmail.com} \\
  \And
  Kushal Raj Bhandari \\
  Rensselaer Polytechnic Institute\\
  Troy, NY \\
  \texttt{bhandk@rpi.edu} \\
  \And
  Jianxi Gao \\
  Rensselaer Polytechnic Institute\\
  Troy, NY \\
  \texttt{gaoj8@rpi.edu} \\
}

\begin{document}

\maketitle

\begin{abstract}
%leave it  blank
Garbage production and littering are persistent global issues that pose significant environmental challenges. Despite large-scale efforts to manage waste through collection and sorting, existing approaches remain inefficient, leading to inadequate recycling and disposal. Therefore, developing advanced AI-based systems is less labor intensive approach for addressing the growing waste problem more effectively. These models can be applied to sorting systems or possibly waste collection robots that may produced in the future. AI models have grown significantly at identifying objects through object detection.This paper reviews the implementation of AI models for classifying trash through object detection, specifically focusing on the use of YOLO V5 for training and testing. The study demonstrates how YOLO V5 can effectively identify various types of waste, including \textit{plastic}, \textit{paper}, \textit{glass}, \textit{metal}, \textit{cardboard}, and \textit{biodegradables} \footnote{\href{https://github.com/everestkuang/GarbageDetectionYoloV5}{Github: GarbageDetectionYOLOV5} }.

\end{abstract}

\section{Introduction}
%start writing here
Garbage waste has been an enduring issue in the world. A total of 2.12 billion tons of waste is produced each year[6]. This problem is prevalent everywhere—in oceans, rivers, forests, and streets—where one can easily spot plastic bottles, cigarette butts, metal cans, and other types of trash. Such waste has nearly reached every corner of this Earth. The World Bank reports that global waste is expected to increase by 70 percent in the year 2050[1], which computes to an astounding 3.6 billion tons. With the growing production and consumption of goods, garbage will remain an ever-present constant in the world. A solution to this global problem is the collection, categorization, and then systematic management of different types of solid waste. 

However, even with the collection of garbage to protect the environment, a question still remains: What are we to do with the garbage? Managing this amount of waste is a great undertaking, and much care is needed to successfully make good use of the garbage that we have. As of now, The World Bank states that at least 33 percent of trash is mishandled [10]. Oftentimes, they are immediately burned or sent to landfills. In order to efficiently reuse trash, it requires a system that can sort the waste into different categories. For humans, it is cost and time-inefficient to spend countless hours sorting through waste efficiently. 

This is where an autonomous system can be used to efficiently sort garbage into different categories based on what material they are made of. With the use of AI, collecting and sorting garbage will become extremely easy, and this will allow us to not heap all the garbage into wastelands or burn it, causing more emissions of toxic or unhealthy gases. I see the potential possibility of an autonomous system capable of collecting solid waste in the near future. However, it is still unexplored as to whether the recent advancements in machine learning can be applicable to garbage detection in the real world.

In this paper, I created a You Only Look Once(YOLO) model that uses object detection to classify trash and analyze the performance of the model on real-world test data that can be found in local neighborhoods. I trained the model with an open-source dataset from RoboFlow [2] so that the model can identify and classify waste into different types. Afterwards, I studied the performance of the model on the training data and the real-world data. 

\section{Related Works}

\begin{wrapfigure}{R}{0.5\textwidth} 
  \centering
  \includegraphics[width=0.5\textwidth]{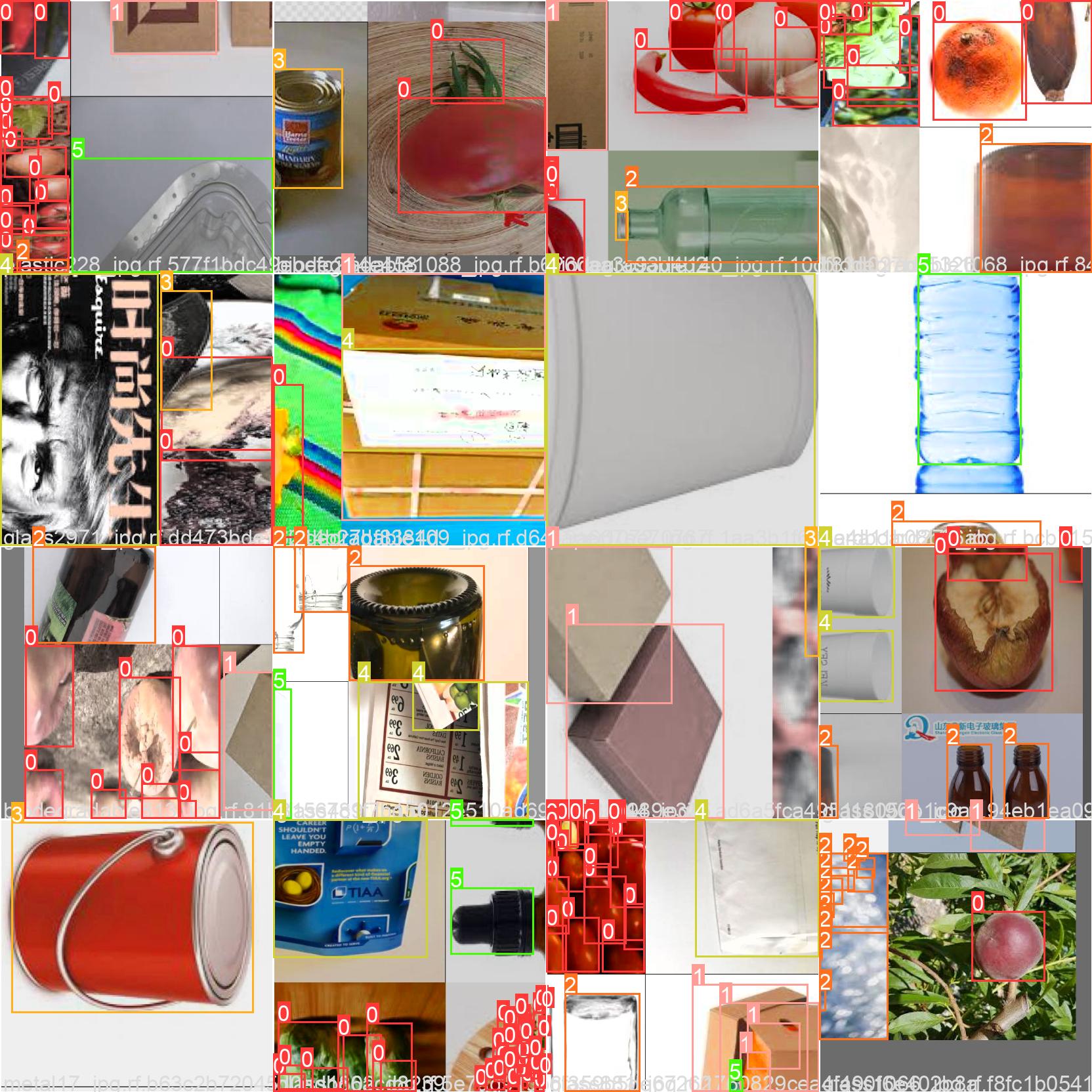}
  \caption{Example of Garbage Detection using YOLO model}
  \label{fig:example_detection}
\end{wrapfigure}

The application of AI models in object detection has been explored with its use in trash classification. 

%https://www.sciencedirect.com/science/article/pii/S0956053X20302105?via%3Dihub 
Piotr Nowakowski and Teresa Pamula made a study on the use of CNN models in classifying e-waste [4]. They addressed the problem with e-waste collection due to its limits by making use of smartphones. The collection of e-waste required lots of planning of routes and collection, so the use of people's smartphones to take pictures of e-waste would be used to help with this planning. A mobile app would be created to classify waste by taking pictures on the phone. The app would use a CNN network to classify the waste and a R-CNN model to detect its category and size. The model's accuracy was found to lie between 90-97 percent, proving to be very effective at classifying e-waste. As a result of the data given by the app, e-waste collection companies would be able to create efficient collection routes and put resources in the right places. 

%https://www.ncbi.nlm.nih.gov/pmc/articles/PMC9412171/ 
In addition to that, Zailan et al. review waste detection of trash in riverine systems through a YOLO model [11]. They address the challenges with waste detection within riverine systems due to the unique conditions that water creates. Pictures of trash from rivers are affected by different illumination levels, image complexity, objects being obscured by water or other objects. However, despite these challenges, the created Yolo model was able to do very well with a 89 percent precision value. This shows the great capabilities of these systems to overcome problems and be very efficient at classifying objects. 

%https://arxiv.org/abs/2208.00773 
Similarly, Viswanatha et al. analyze the difference between the use of YOLO and CNN models in object detection[9]. The YOLO model had greater efficiency, speed, and accuracy since it had a single pass structure and each successive version improved accuracy and speed. The CNN model was better at feature extraction due to its more complex architecture and ability to be customized to certain tasks. It was found that in order for any model to perform well, the system needed to be trained properly with the right datasets. With the comparison of differents there weren't any major differences, however, the YOLO model's speed and accuracy was a key takeaway for their finding. 

\section{Methodology}
%start writing here
\begin{wrapfigure}{R}{0.5\textwidth} 
  \centering
  \includegraphics[width=0.5\textwidth]{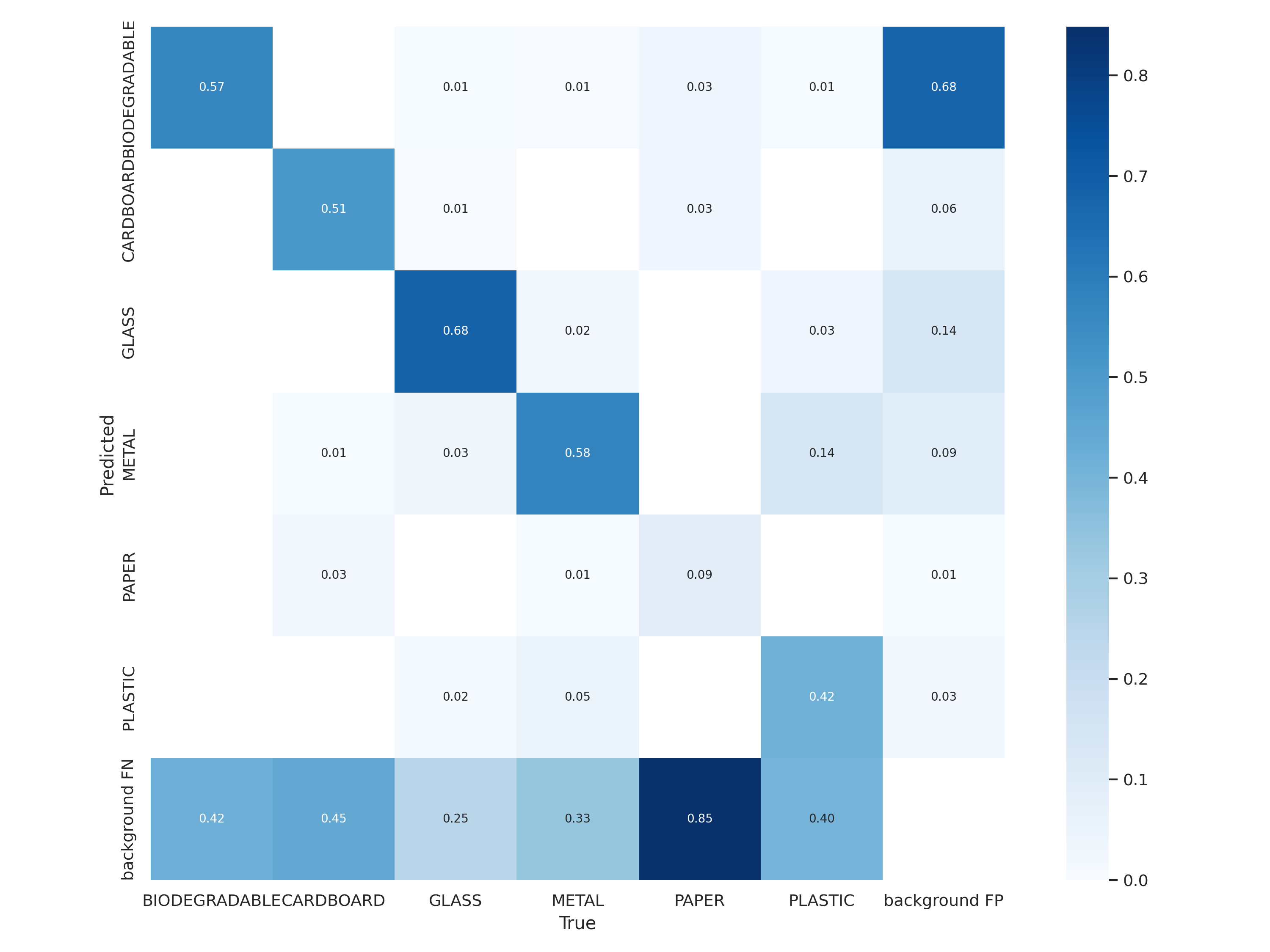}
  \caption{Confusion Matrix for classification of different types of garbage with background}
  \label{fig:confusion_matrix}
\end{wrapfigure}
Recent studies showcase that the YOLO architecture has greater potential for garbage detection rather than any other neural network architecture for garbage detection [9]. Hence, I focus on analyzing the performance of the YOLO model on real-world garbage detection.

The YOLO model is a state-of-the-art object detection algorithm known for its speed and accuracy [5, 9, 11]. I utilize an existing pre-trained YOLO V5 model for object detection from Ultralytics \footnote{https://www.ultralytics.com} [7, 8]. Ultralytics has developed a specific implementation of YOLO (YOLO V5), which includes various enhancements and optimizations. The YOLO V5 model is the free and reliable version of the YOLO model released by Ultralytics. The model is comprised of a backbone, neck, and head. The CSPDarknet53 is the backbone that extracts features from images, and the neck, PANet (Path Aggregation Network), enhances feature fusion over different scales. The head of the model uses anchor-based detection to predict bounding boxes, classes, and confidence scores [8]. The loss function combines the localization loss and classification loss to find class probabilities and bounding box coordinates. The purpose of bounding boxes is to create a rectangle around detected objects so that the location and size of the object can be found, and it is easier to analyze the object.

I utilized the Garbage Classification dataset from Roboflow, an open-source platform, for fine-tuning the pre-trained model. The Garbage Classification dataset contains over 10,000 images to be used for training machine learning models. There are 7234 images in the trainset, 2098 images in the valid set, and 1042 images in the test set. The dataset has 6 labels used to classify garbage: \textit{plastic}, \textit{paper}, \textit{cardboard}, \textit{metal}, \textit{glass}, and \textit{biodegradables}. 

The trained model was applied to real-time object detection by taking photos from the surrounding environment. The pictures were taken from the local neighborhood\footnote{Latham, NY, US} to represent real-world solid waste that aims to sample garbage distribution around anywhere. Over 100 pictures were gathered, and many had different perspectives, amounts of trash, and environments to see the model's effectiveness at correctly classifying trash.
%Drawback of YOLO model
%However, there were some problems due to the limited number of labels in the model. Some trash in the photos didn't fall under a certain category of class that the model had. For improvements, it would be important for the system to have more labels. Yet, the model would require a better dataset of greater quantity and quality, so that it learns to classify objects correctly. 

%Overall, this YOLO model will be very useful in the future for object detection. It could be used in many implemented in garbage collection to sort garbage on conveyor belts. However, in the future, it could be implemented into robots to aid them in the collection and sorting of such trash. As a result, we could have more efficient cleaning of the environment. 

\section{Result and Analysis}

% Your analysis of the text below
I evaluated the performance of the model during training using validation data while training for 100 epochs. The confusion matrix, Figure \ref{fig:confusion_matrix} compares label predictions to the actual label of the image. The accuracy in the confusion matrix shows that the model performed decently. Each label class did pretty well, with most of them being above 0.5 for correctly predicted labels. For most labels, the model is getting confused with the background. For example, \textit{Biodegradable} and \textit{Paper} waste are being misclassified as background. This may be due to the overlapping of the background over the object itself or objects covering each other. The overlapping of objects and background makes the model to have a harder time in identifying what the object may be.

%\begin{wrapfigure}{L}{0.4\textwidth} 
%  \centering
%  \includegraphics[width=0.4\textwidth]{figure/test_analysis_figure/garbage_type_distribution.png}
%  \caption{Garbage Type Distribution for Real World Test data}
%  \label{fig:garbage_type_distribution}
%\end{wrapfigure}

% Your analysis of the text below
For the testing of the trained model, I collected pictures from various locations with the objective of sampling solid waste in its natural environment. These pictures included a variety of trash in the real world, including metal scraps, plastic bottles, food scraps, and other trash items. While taking pictures, I kept in mind to take as many as possible and try to get a wide range of trash, so that the Yolo model could be tested to its fullest extent. 

\begin{figure}
\begin{minipage}[c]{0.49\linewidth}
  \includegraphics[width=\textwidth]{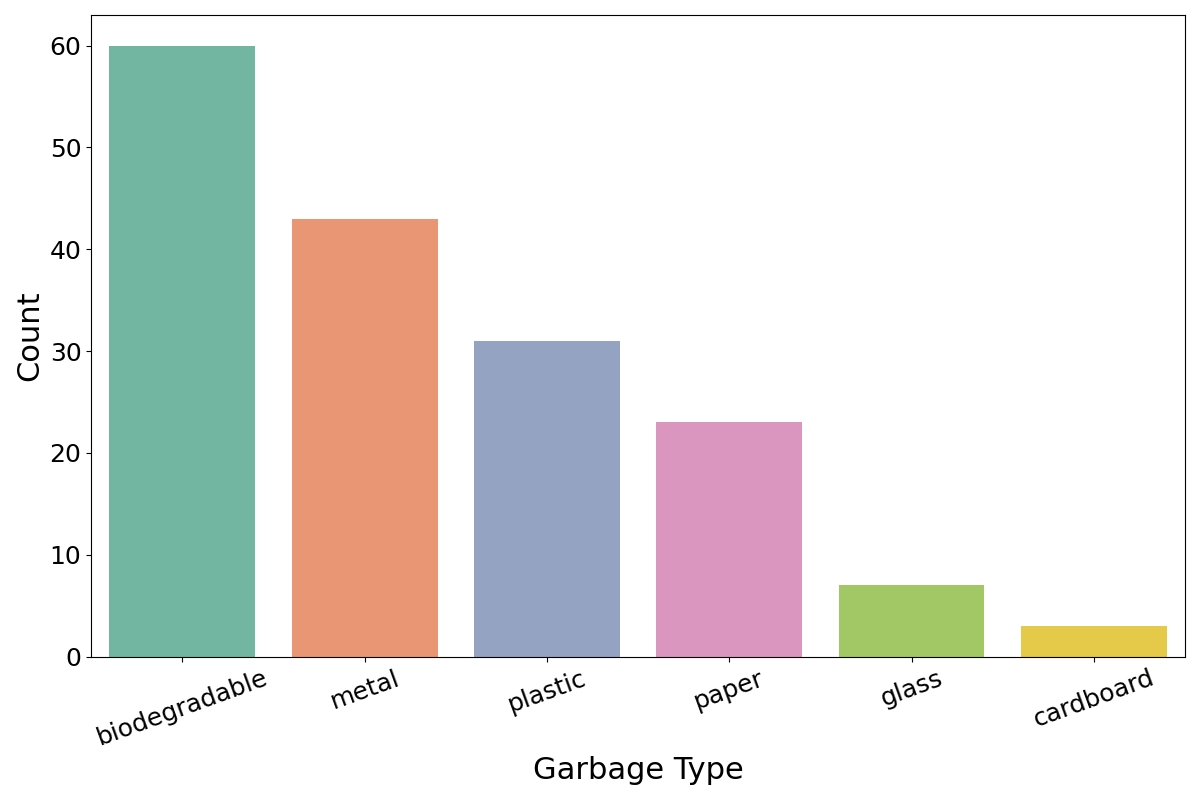}
  \caption{Garbage Type Distribution for Real World Test data}
  \label{fig:garbage_type_distribution}
\end{minipage}
\hfill
\begin{minipage}[c]{0.49\linewidth}
\includegraphics[width=\textwidth]{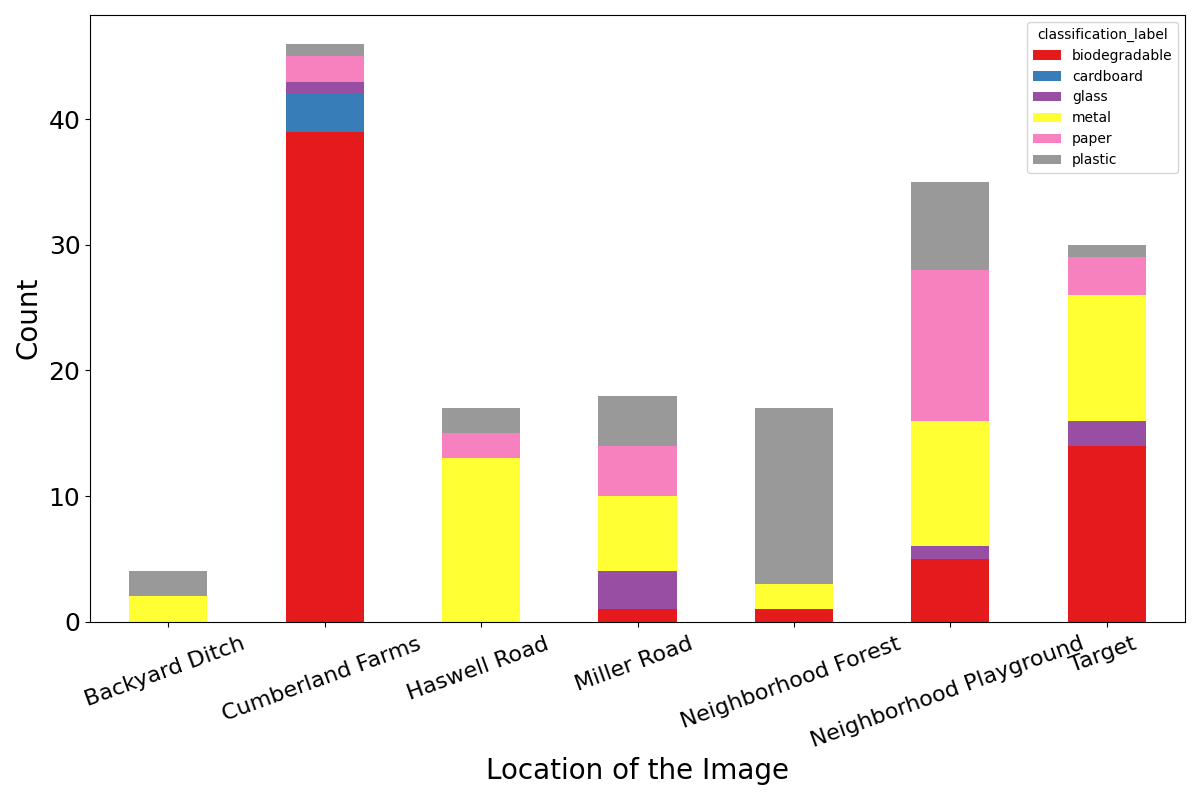}
\caption{Frequency of Type of Garbage based on different type of location}
\label{fig:location_based_distribution}
\end{minipage}%
\end{figure}

In Figure \ref{fig:garbage_type_distribution}, the graph shows the frequency of the type of garbage. The model classifies \textit{Biodegradables} as the most prevalent type of garbage and \textit{Cardboard} as the least prevalent. This resembles the performance with the validation dataset, where most of the misclassifications are from the background being classified as \textit{biodegradable}. This is also connected to the greater amount of objects that resembled biodegradables in these pictures such as leaves, grass, sticks, and other organic matter. 

%\begin{wrapfigure}{L}{0.4\textwidth} 
%%  \centering
%  \includegraphics[width=0.4\textwidth]{figure/test_analysis_figure/location_based_distribution.png}
%  \caption{Frequency of Type of Garbage based on different type of location}
%  \label{fig:location_based_distribution}
  
%\end{wrapfigure}
% Your analysis of the text below

Prior to testing, I classified images based on the location of the solid waste. These places included local stores, roads, and neighborhood areas. Due to the different amounts of human activity in these areas, there was a greater amount of trash near certain areas, such as stores or playgrounds. We can see from Figure \ref{fig:location_based_distribution} certain trends in different areas with the different types of trash found in certain areas, as seen by the greater amount of \textit{biodegradables} at Cumberland Farms or the large amount of \textit{plastic} in the neighborhood forest. 

\begin{wrapfigure}{R}{0.5\textwidth} 
  \centering
  \includegraphics[width=0.5\textwidth]{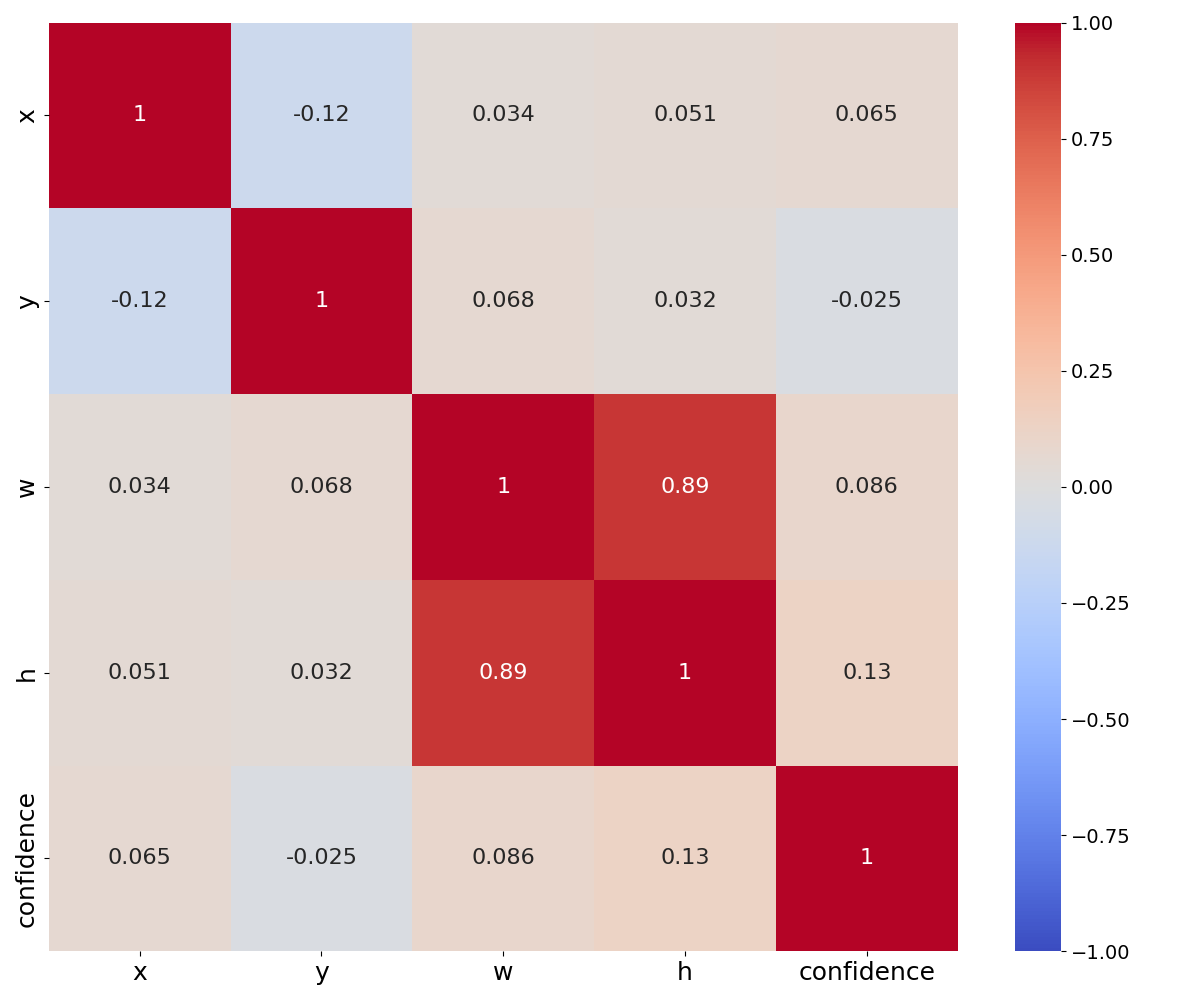}
  \caption{Correlation Matrix of a different dimension of the object recognized with the confidence level}
  \label{fig:correlation_matrix}
\end{wrapfigure}

% Your analysis of the text below
The correlation matrix, Figure \ref{fig:correlation_matrix}, compares the relationship between multiple variables in the image. Increased intensity shows a relationship between two variables, such as the correlation between width and height variables. This indicates that wider bounding boxes will generally be taller. Most variables have little correlation with other variables, with the exception of the width and height variables. 
%reflection
\section{Conclusion}
%start writing here
 From this research paper, a YOLO model can be created to classify trash into multiple categories.The pre-trained YOLO V5 model was finetuned for 100 epochs to be tested with pictures from the natural environment. The model was able to classify different types of waste in different categorizes. The results of the study showcase the performance of the model substantially limited for practical usecase. The model, in particular, exhibited certain limitations. The result show that it struggles with complex patterns such as overlapping objects along with limitations in detecting objects, mainly when the background is not simple. However, some of these problems could be resolved by utilizing complex type of garbage classification dataset for training the model. Some real world waste in the photos didn't fall under any certain category of label. For example, some photos included rubber, polyester, or other materials. This would require the model to be trained on more diverse labels. The result show case that more work can be done to improve the system for real world application. 
 
 Despite these shortcomings, YOLO model have great potential to be used for trash classification in helping collection and disposal. With the growing progress of generative AI, like diffusion model, the existing model could be improved by isolating complex background and make use of existing dataset for better object detection and classification. AI models clearly display great potential in helping to manage trash, as seen by their ability to transcend just classifying trash and provide trends on the garbage types at certain locations. This can help in fighting trash waste by looking at waste patterns of type and area. In conclusion, AI models can have a big impact on waste management, and they can be improved further to be utilized properly and more efficiently.

\newpage

\begin{ack}
In this research, Kushal Bhandari has been a huge help for me in teaching me everything about AI models from CNN models to Yolo models. He was the only advisor I had for helping me to learn about AI and machine learning. For this paper, he has given me some guidance and reviewed my paper. He supervised all my research and work, and he helped with finalizing my paper. 

I recieved no funding for this work.
\end{ack}

\section*{References}
\small{
%put citation here
[1] “Did You Know? In 2050, the World Will Generate 3.4 Billion Tonnes of Waste per Year.” Landfillsolutions, 9 Nov. 2021, landfillsolutions.eu/did-you-know-the-world-will-generate-3-4-billion-tonnes-of-waste-per-year-in-2050/.

[2] Identification, Material. Roboflow, 26 Mar. 2022, universe.roboflow.com/material-identification/garbage-classification-3.

[3] “National Overview: Facts and Figures on Materials, Wastes and Recycling | US EPA.” US EPA, 22 Nov. 2023, www.epa.gov/facts-and-figures-about-materials-waste-and-recycling/national-overview-facts-and-figures-materials.

[4] Nowakowski, Piotr, and Teresa Pamuła. “Application of Deep Learning Object Classifier to Improve E-waste Collection Planning.” Waste Management, vol. 109, May 2020, pp. 1–9. https://doi.org/10.1016/j.wasman.2020.04.041.

[5] Redmon, Joseph, et al. “You Only Look Once: Unified, Real-Time Object Detection.” arXiv.org, 8 June 2015, arxiv.org/abs/1506.02640.

[6] The World Counts. www.theworldcounts.com/challenges/planet-earth/state-of-the-planet/world-waste-facts.

[7] Ultralytics. “Comprehensive Guide to Ultralytics YOLOv5.” Ultralytics YOLO Docs, 22 June 2024, docs.ultralytics.com/yolov5.

[8] Ultralytics. “GitHub - Ultralytics/Yolov5: YOLOv5 in PyTorch > ONNX > CoreML > TFLite.” GitHub, github.com/ultralytics/yolov5.

[9] V, Viswanatha, et al. “Real Time Object Detection System With YOLO and CNN Models: A Review.” arXiv.org, 23 July 2022, arxiv.org/abs/2208.00773.

[10] What a Waste. datatopics.worldbank.org/what-a-waste.

[11] Zailan, Nur Athirah, et al. “An Automated Solid Waste Detection Using the Optimized YOLO Model for Riverine Management.” Frontiers in Public Health, vol. 10, Aug. 2022, https://doi.org/10.3389/fpubh.2022.907280.
}
\appendix
%start writing here
\section{Appendix: Training Dataset}

%===================== These are Labels and x and y coordinate ===================

\begin{figure}[ht]
  \centering
  \includegraphics[width=0.9\textwidth]{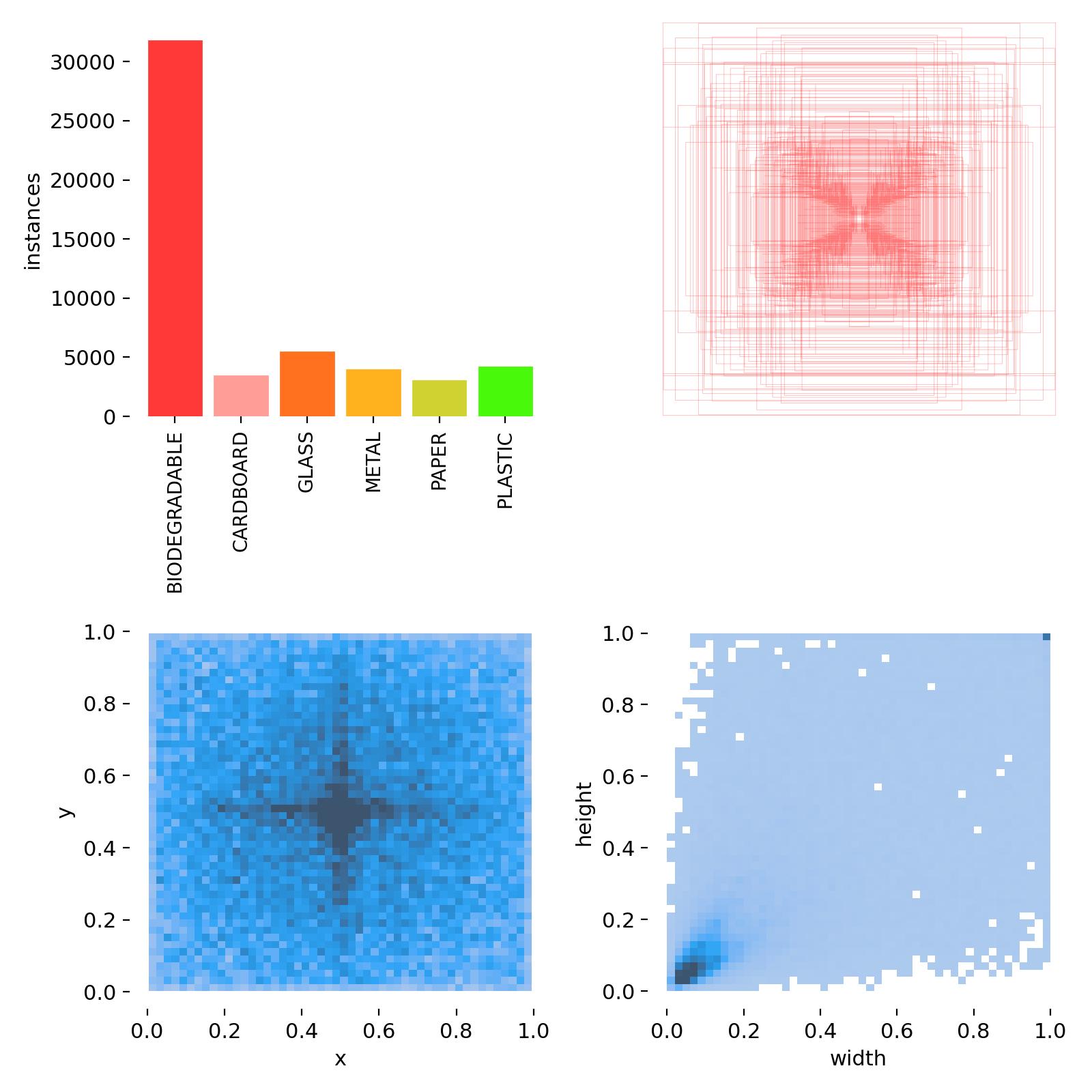}
  \caption{Statistics of object detection labels, classification labels and object detection labels}
  \label{fig:train_labels}
\end{figure}

% Your analysis of the text below
In the training batch, Figure \ref{fig:train_labels} plots were created for data on object detection labels and bounding boxes. The first chart described the instances of each label class, and the second gave an overlapping rectangular plot for regions of interest in images. The bottom two plots were density graphs comparing the place of data points with x and y coordinates and height and width.

The first chart to the top left gives data on the number of instances for certain label classes. Each waste type was given a label class, and different amounts of each were trained. Generally, there were an equal number of instances for every class, except for \textit{Biodegradables}, for which there was a greater number in the training dataset. 

The plot in the top right shows the regions of interest in each image and where bounding boxes were generally created. The graph shows a general concentration in the middle for all images, meaning that in most images trash was generally in the middle with some scattered to the sides. 

The x and y density graph illustrates the distribution of data points. There was a greater amount of density in the middle, agreeing with the overlapping rectangular chart above where there was a greater amount of bounding boxes in the middle. As seen by the density, most of the data points were clustered in the middle of the plot.  

The second density graph compares the height and width of data points. Most data points seemed to have a very small height and width seen by the greater density in the bottom left corner. 

\section{Appendix: Performance with the Validation Data}
%===================== These are F1 and P_Curve for Training Validation data ===================
\begin{figure}[ht]
  \centering
  \includegraphics[width=0.5\textwidth]{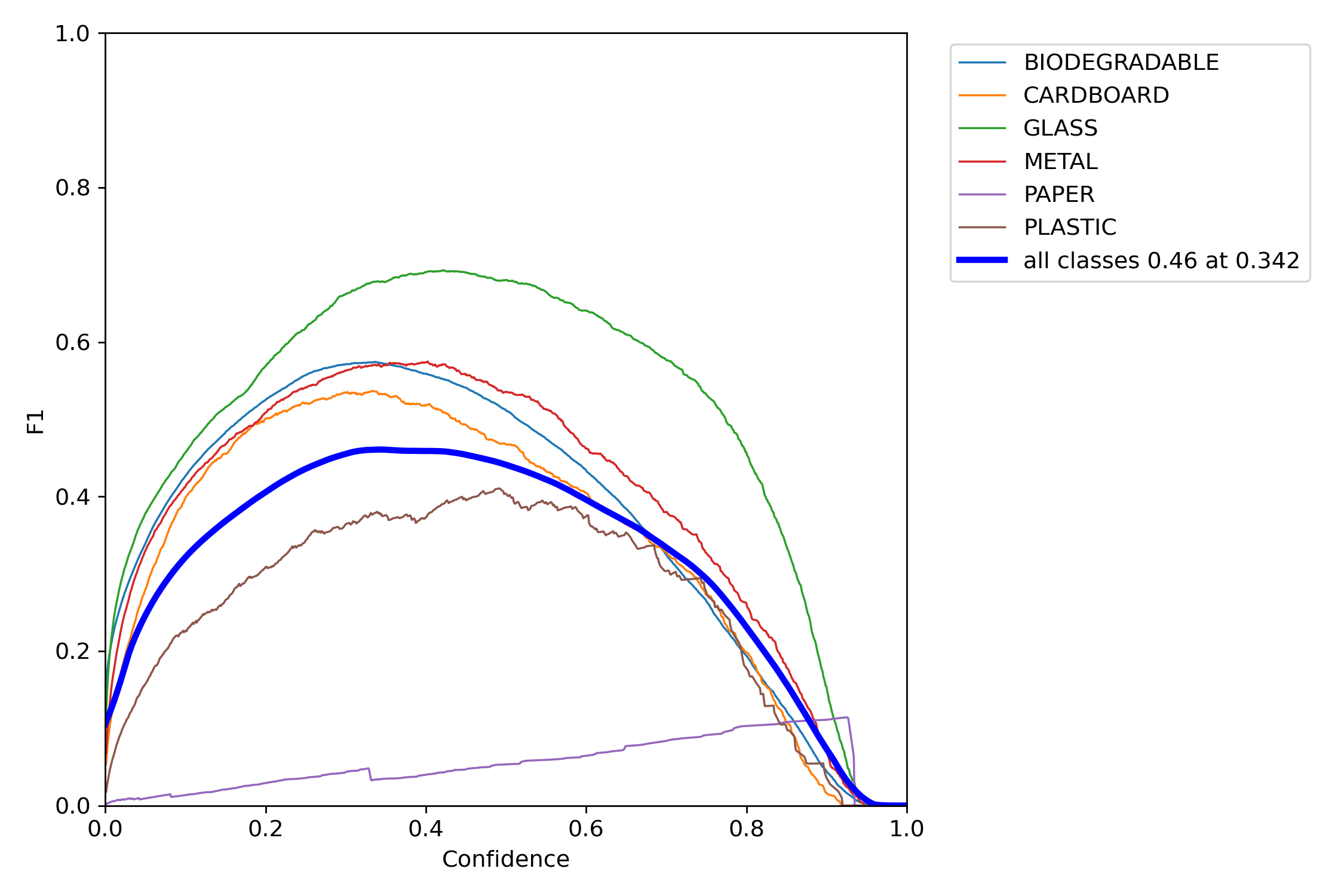}\hfill
  \includegraphics[width=0.5\textwidth]{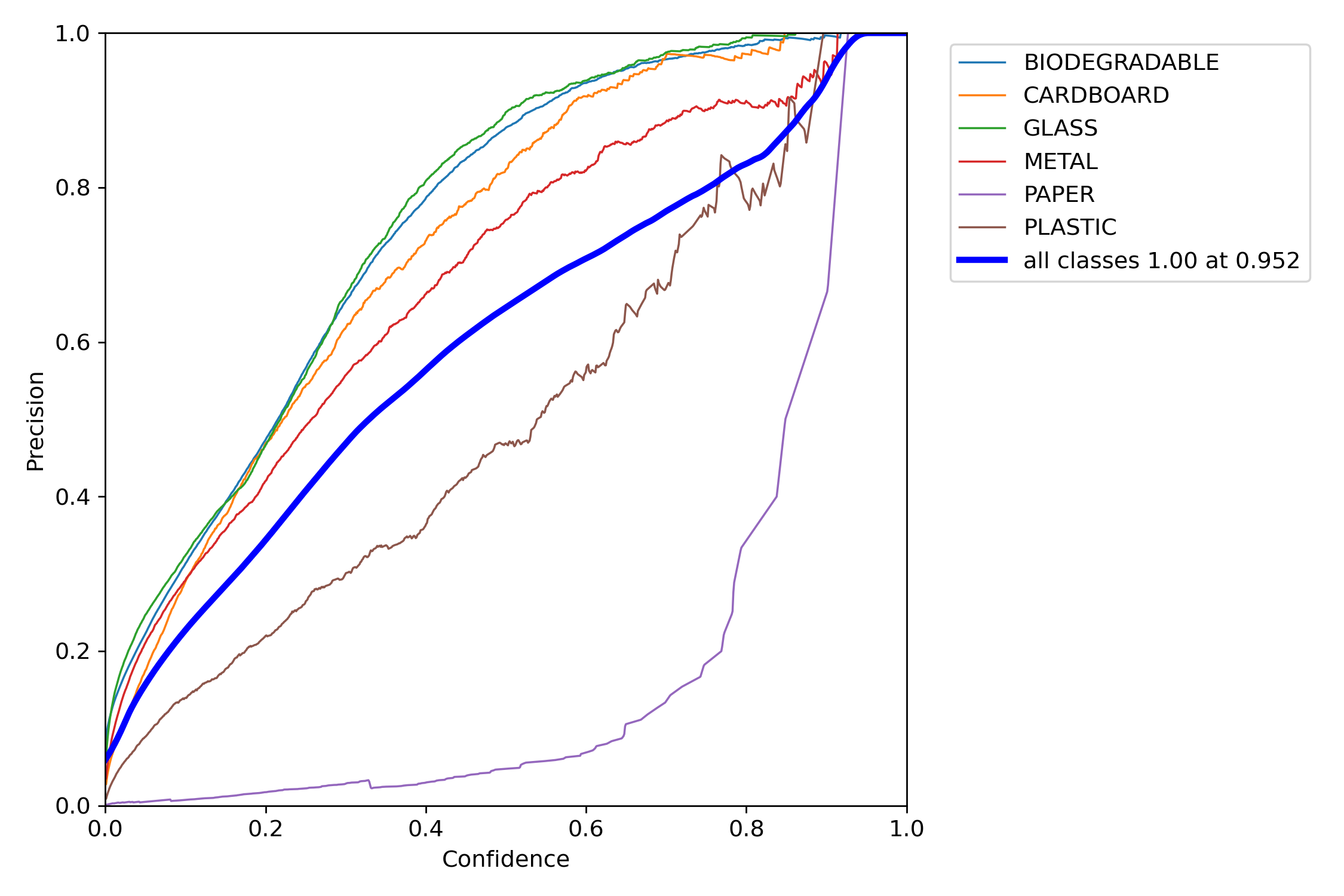}
  \caption{F1 score and Precision of different labels over confidence level}
  \label{fig:f1_score_vs_confidence}
\end{figure}
% Your analysis of the text below
 With Figure \ref{fig:f1_score_vs_confidence}, the confidence level was compared with the model's F1 score and precision of different labels. F1 score specifically measures the accuracy using precision and recall specific to that certain label class. Precision relates to the successful predictions made by the model.

On the left, the graph comparing confidence F1 scores to confidence levels shows the model's predictive ability. The model performed its best at roughly 0.2 to 0.6 confidence score level. \textit{Plastic} did very poorly, barely rising over 0.1 for an F1 score at its max, and did not follow the general trend compared to other label classes. 

For the comparison between precision scores and confidence levels, we saw a general increase in precision as confidence levels went up. Many class labels generally increase accuracy quickly as confidence scores go up. However, the model struggles with \textit{Plastic} and \textit{Paper} with greater confidence. As seen in the confusion matrix, \textit{Paper} was especially hard for the YOLO model to classify correctly. 

\section{Appendix: Analysis of Confidence Level on Test Data}
\begin{figure}[ht]
\begin{subfigure}[h]{0.49\linewidth}
\includegraphics[width=\linewidth]{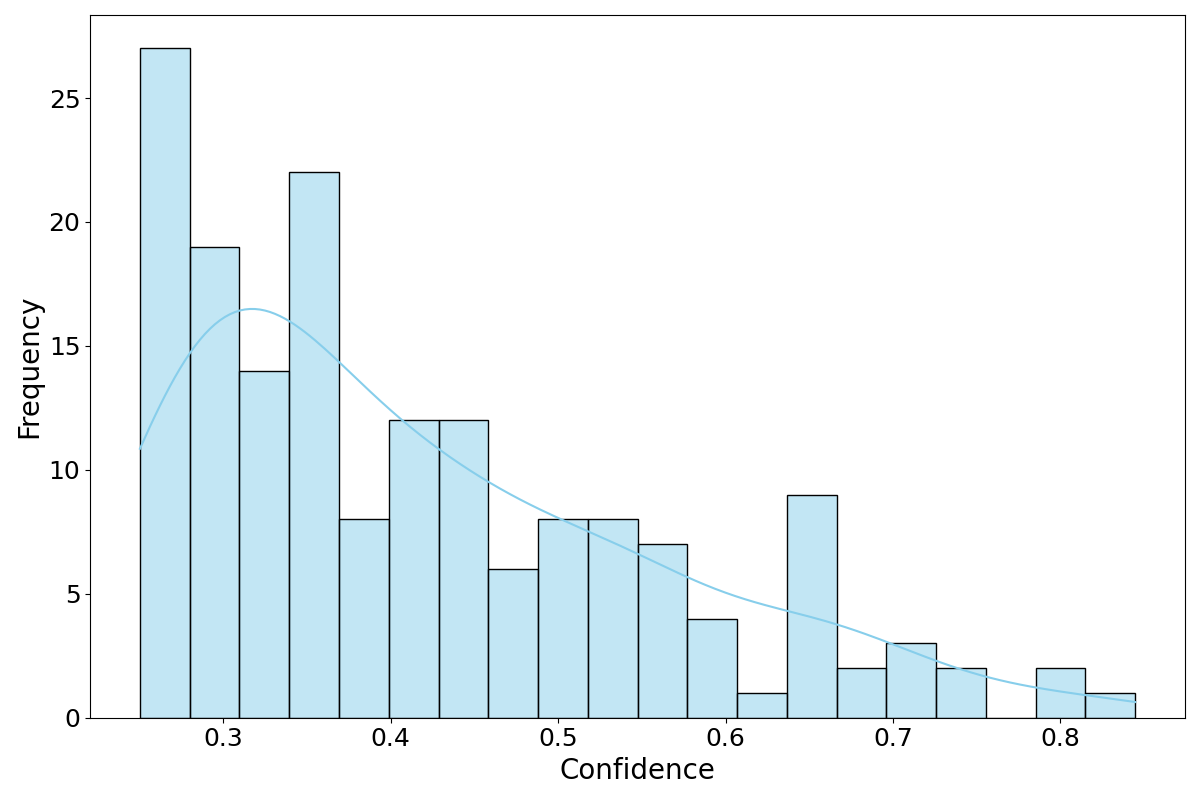}
\caption{Frequency of Confidence Level for trained model on Real World Data}
\end{subfigure}
\hfill
\begin{subfigure}[h]{0.49\linewidth}
\includegraphics[width=\linewidth]{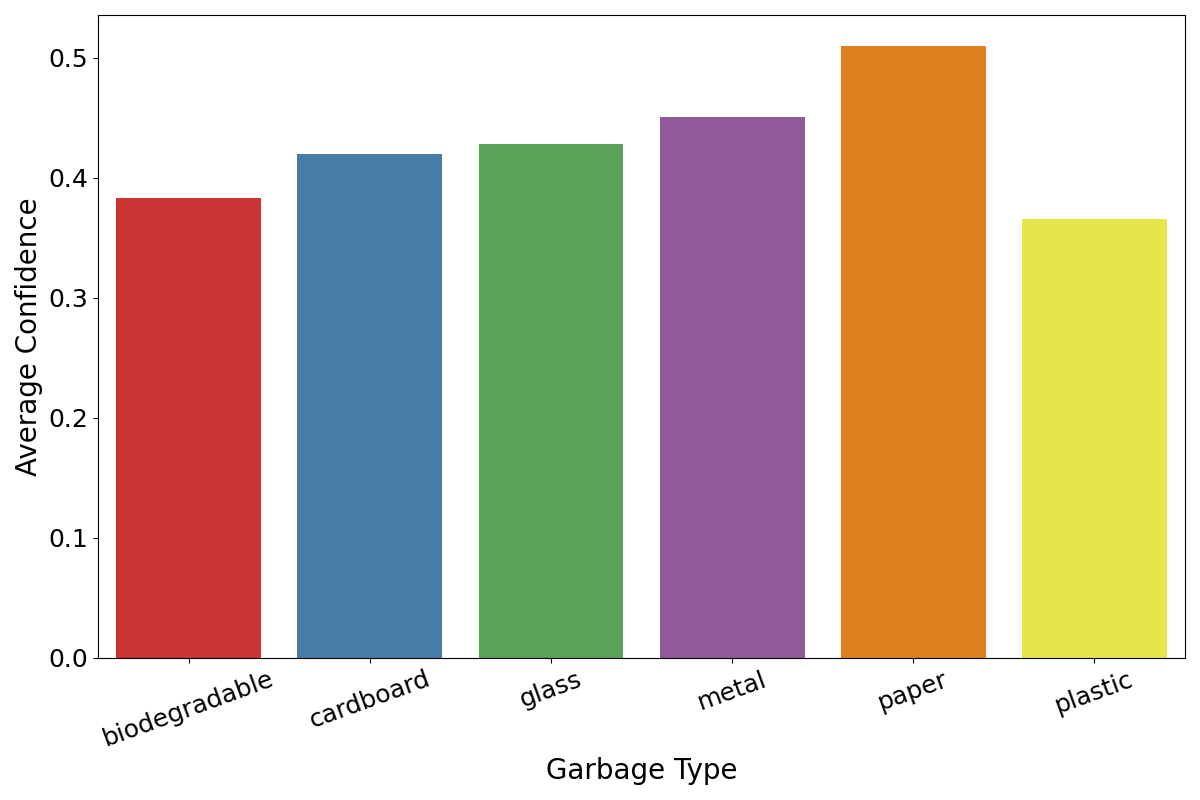}
\caption{Average Confidence Level to different types of garbage detected}
\end{subfigure}%
\caption{Confidence Level of the trained model on Real World Data}
\label{fig:confidence_level_freqeuncy}

\end{figure}

% Your analysis of the text below
The bar chart in Figure \ref{fig:confidence_level_freqeuncy} a. displays the frequency of different confidence levels on the test data. There is a greater amount of smaller confidence levels seen by the higher frequency of confidence levels in the lower range compared to that of the higher range. Generally, at a confidence level of 0.3, there were 20+ instances of this confidence level. This is a much greater number than the confidence level of 0.8, which had less than 5 instances. The curve on the graph shows the decreased frequency for higher confidence levels. 

Similarly, when the confidence levels for garbage types were compared, Figure \ref{fig:confidence_level_freqeuncy} b., they were all around the 0.3 to 0.5 range. The model's lack of confidence levels was due to problems with the image, as light intensity, blurriness, and obscuring of items affected its ability to classify waste properly. Another contributing factor could be the background of the test images were real-world images in comparison to the test dataset, where the background is much simpler.

\end{document}